%% file: main.tex
\documentclass[11pt,letterpaper]{article}

\usepackage{PRIMEarxiv}

\usepackage[utf8]{inputenc} 
\usepackage[T1]{fontenc}    
\usepackage{hyperref}       
\usepackage{url}            
\usepackage{booktabs}       
\usepackage{amsfonts}       
\usepackage{nicefrac}       
\usepackage{microtype}      
\usepackage{lipsum}
\usepackage{fancyhdr}       
\usepackage{graphicx}       
\usepackage{subfiles}
\usepackage{xcolor}
\usepackage{multirow}
\usepackage{enumerate}
\usepackage{subfiles}
\usepackage{multirow}
\usepackage{graphicx}
\usepackage{subfiles}
\usepackage{listings}

\graphicspath{{media/}}     

\newcommand{\sfeat}{(f, \mathbf{P}, \mathbf{L})}
\newcommand{\fpl}{f_{\mathbf{P}\mathbf{L}}}
\newcommand{\mbb}[1]{\mathbb{#1}}
\newcommand{\mbf}[1]{\mathbf{#1}}

\title{Towards White Box Deep Learning
}

\author{
  Maciej Satkiewicz \\
  314 Foundation \\
  Warsaw/Krakow \\
  \texttt{maciej.satkiewicz@314.foundation} \\
}

\begin{document}
\maketitle

\begin{abstract}
Deep neural networks learn fragile "shortcut" features, rendering them difficult to interpret (black box) and vulnerable to adversarial attacks. This paper proposes \emph{semantic features} as a general architectural solution to this problem. The main idea is to make features locality-sensitive in the adequate semantic topology of the domain, thus introducing a strong regularization. The proof of concept network is lightweight, inherently interpretable and achieves almost human-level \textbf{adversarial} test metrics - with \textbf{no adversarial training}! These results and the general nature of the approach warrant further research on semantic features. The code is available at \url{https://github.com/314-Foundation/white-box-nn}.
\end{abstract}

\section{Introduction}\label{sec:introduction}
\input{introduction}

\section{Related work}\label{sec:related}
\input{related_work}

\section{Methodology}\label{sec:methodology}
\input{methodology}

\section{Semantic features}\label{sec:theory}
\input{theory}

\section{Network architecture}\label{sec:architecture}
\input{architecture}

\section{Results}\label{sec:results}
\input{results}

\section{Discussion}\label{sec:discussion}
\input{discussion}

\section{Further research}\label{sec:further}
\input{further_research}

\section{Acknowledgments}\label{sec:acknowledgments}
\input{acknowledgments}

\newpage
\bibliographystyle{plainurl}
\bibliography{references}  

\appendix
\newpage

\input{appendix}

\end{document}

%% file: introduction.tex
The main advantages of deep neural networks (DNNs) are their architectural simplicity and automatic feature learning. The latter is crucial for working with unstructured data as developers don't need to design features by hand. However, giving away the control over features leads to \emph{black box} models - \textbf{DNNs tend to learn hardly interpretable "shortcut" correlations~\cite{Geirhos_2020} that leak from train to test~\cite{ilyas2019adversarial}}, hampering alignment and out-of-distribution performance. In particular, this gives rise to adversarial attacks~\cite{szegedy2014intriguing} - semantically negligible perturbations of data that arbitrarily change model's predictions. Adversarial vulnerability is a widespread phenomenon (vision~\cite{szegedy2014intriguing}, segmentation/detection~\cite{xie2017adversarial}, speech recognition~\cite{carlini2018audio}, tabular data~\cite{cartella2021adversarial}, RL~\cite{gleave2021adversarial}, NLP~\cite{zou2023universal}) and largely contributes to the general lack of trust in DNNs, substantially limiting their adoption in high-stakes applications such as healthcare, military, autonomous vehicles or cybersecurity.
Conversely, the main advantage of hand-designed features is the fine-grained control over model's performance; however, such systems quickly become infeasibly complex.

\begin{figure}
  \centering
  \includegraphics[scale=0.35]{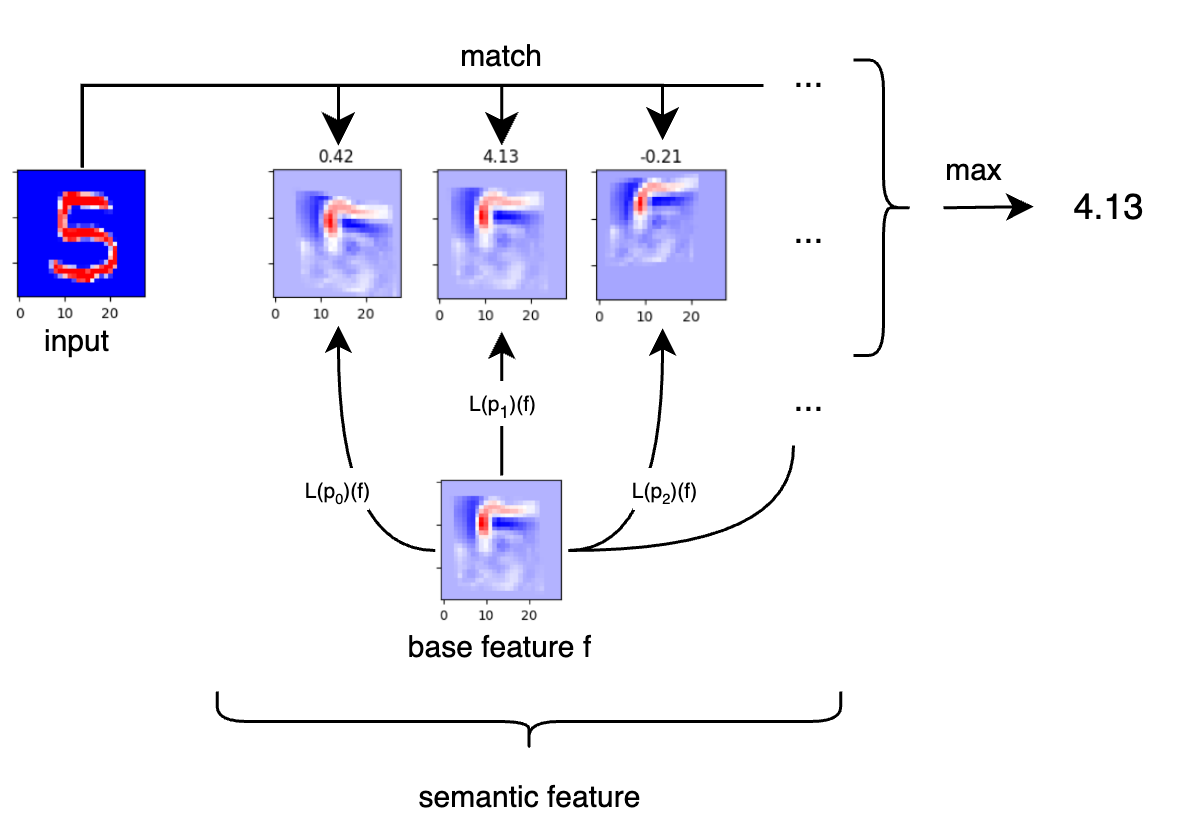}
  \caption{Visualisation of \emph{semantic feature} $\fpl$ and its matching mechanism (SFmatch, see Section~\ref{sec:SFmatch}). The semantic feature $\fpl$ consists of a \emph{base feature} $f$ and a set of its "small" modifications $\mathbf{L}(p_i)(f)$. In this case, the base feature has the same dimensions as the input image and its modifications are 2D affine transformations. $\mathrm{SFmatch}(d, \fpl)$ takes the maximum of the scalar products $d\cdot\mathbf{L}(p_i)(f)$ thus identifying all $\mathbf{L}(p_i)(f)$. Affine transformations are typically parameterized by extended matrices of dimension $2\times3$; both the base feature and the parameters $p_i$ of its modifications are \textbf{learned}. If $f$ and its modifications are easily understood by humans, then the neural network layer composed of such features can be considered a \emph{white box} layer. The precise definition of semantic feature is found in Section~\ref{sec:theory}.}
  \label{fig:SFmatch}
\end{figure}

This paper aims to address those issues by reconciling Deep Learning with feature engineering - with the help of \emph{locality engineering}. Specifically, \emph{semantic features} are introduced as a general conceptual machinery for controlled dimensionality reduction inside a neural network layer. Figure~\ref{fig:SFmatch} presents the core idea behind the notion and the rigorous definition is given in Section~\ref{sec:theory}. Implementing a semantic feature predominantly involves encoding appropriate invariants (i.e. semantic locality) for the desired semantic entities, e.g. "small affine perturbations" for shapes or "adding small $\delta\in[-\epsilon, \epsilon]$" for numbers. Thus, semantic features capture the core characteristic of real-world objects: \emph{having many possible states but being at exactly one state at a time}. It turns out that for the trained proof-of-concept (PoC) DNN such parameter-sharing strategy leads to human-aligned \emph{base features} - which motivates calling the PoC a \emph{white box} neural network, see Section~\ref{sec:results}.
In particular, the resulting inherent interpretability of the PoC allowed for designing a strong \textbf{architectural defense}\footnote{with no form of adversarial training!} against adversarial attacks, as discussed in Section~\ref{sec:discussion}. Thus, the notion of "semantic feature" and the adversarially robust white box PoC network are the main contributions of this work.

The paper is organised as follows. Section~\ref{sec:related} investigates related work on explainability and adversarial robustness. Section~\ref{sec:methodology} briefly describes the adopted methodology. Section~\ref{sec:theory} gives the rigorous definition of \emph{semantic feature} and argues for the generality of the notion. Section~\ref{sec:architecture} builds a carefully motivated PoC 4-layer white box neural network for the selected problem. Section~\ref{sec:results} analyses the trained model both qualitatively and quantitatively in terms of explainability, reliability, adversarial robustness and performs an ablation study. Section~\ref{sec:discussion} provides a short discussion of advantages and limitations of the approach. The paper ends with ideas for further research and acknowledgments.

%% file: related_work.tex
\subsection{Adversarial robustness}

Existing methods of improving DNN robustness are far from ideal. Adversarial Training~\cite{madry2019deep} being one of the strongest \emph{empirical} defenses against attacks~\cite{athalye2018obfuscated} is slow, data-hungry~\cite{schmidt2018adversarially, li2022robust} and results in robust generalisation gap~\cite{rice2020overfitting, yu2022understanding} as models overfit to the train-time adversaries. This highlights the biggest challenge of empirical approaches, i.e. reliable evaluation of robustness, as more and more sophisticated attacks break the previously successful defenses~\cite{carlini2019evaluating, croce2020reliable}.

To address this issue the field of \emph{certified robustness} has emerged with the goal of providing theoretically certified upper bounds for adversarial error~\cite{li2023sok}. However the complete certification faces serious obstacles as it is NP-Complete~\cite{weng2018fast} and therefore limited to small datasets and networks, and/or to probabilistic approaches. Another shortcoming is that such guarantees can only be given for a precisely defined \emph{threat model} which is usually assumed to be a $L_p$-bounded perturbation. The methods, results and theoretical obstacles can be vastly different\footnote{for example SOTA certified robustness for MNIST is \textasciitilde94\% for $L_{\infty}, eps=0.3$ and only \textasciitilde73\% for $L_2, eps=1.58$, see \url{https://sokcertifiedrobustness.github.io/leaderboard/}.} for different $p \in \{0, 1, 2, \ldots, \infty\}$ and often require training the model specifically towards the chosen $L_p$-certification.

\subsection{Explainability}

Recent work indicates a link between adversarial robustness and inherent interpretability as adversarially trained models tend to enjoy Perceptually-Aligned Gradients (PAG) - taking gradient steps along the input to maximise the activation of a selected neuron in the representation space often yields visually aligned features~\cite{engstrom2019adversarial, kaur2019perceptuallyaligned}. There is also some evidence for the opposite implication that PAG implies certain form of adversarial robustness~\cite{ganz2023perceptually, srinivas2024models}. However, Section~\ref{sec:ablation} of this paper shows that the latter implication does not hold - model can be inherently interpretable but not robust.

Susceptibility to adversarial attacks is arguably the main drawback of known approaches to Explainable AI (XAI). One would reasonably expect explainable models to rely exclusively on human-aligned features instead of spurious "shortcut" signals~\cite{Geirhos_2020} and therefore be robust to adversaries. However, the \emph{post-hoc} attribution methods, which aim to explain the decisions of already-trained models, are fragile~\cite{adebayo2020sanity} and themselves prone to attacks~\cite{ghorbani2018interpretation}. On the other hand, the \emph{inherently interpretable} solutions such as Prototypical Parts \cite{li2017deep, chen2019looks}, Bag-of-Features~\cite{brendel2019approximating}, Self-Explaining Networks~\cite{alvarezmelis2018robust}, SITE~\cite{wang2021selfinterpretable}, Recurrent Attention~\cite{8099959}, Neurosymbolic AI~\cite{sheth2023neurosymbolic} or Capsule Networks~\cite{sabour2017dynamic} all employ various classic black box encoder networks and thus introduce the adversarial vulnerability into their cores.

Overall the field of adversarially robust explanations (AdvXAI) is fairly unexplored; for more information on this emerging research domain see the recent survey~\cite{Baniecki_2024}.

\subsection{Geometric Deep Learning}

The idea behind semantic features resembles the framework of Geometric Deep Learning~\cite{bronstein2021geometric} or the more recent Categorical Deep Learning~\cite{gavranović2024categorical} in that semantic features are invariant to small deformations (local symmetries). However, semantic features are more practical as they focus on the local invariance and don't require defining a formal structure on the input space; instead, they \textbf{learn to sample} the semantic neighbourhood of a \emph{base feature} (which is also learned). Thus, the only effort required from the developer is to implement sufficient \emph{sampling mechanism}, which doesn't need to be formal or exhaustive, e.g. small affine perturbations seem to be good enough for MNIST features (see Section~\ref{sec:affine_results}), despite the fact that not every possible local perturbation of these features is affine. Additionally, the mentioned frameworks are high-level approaches that don't address explicitly the core issue of adversarial vulnerability.

%% file: methodology.tex
\begin{figure}
  \centering
  \includegraphics[scale=0.5]{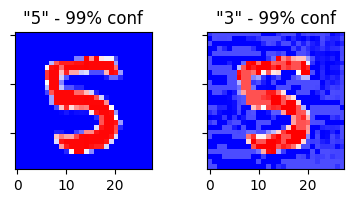}
  \caption{A typical DNN for MNIST can be arbitrarily fooled by adding semantically negligible noise.}
  \label{fig:mnist_3_5_adv}
\end{figure}

This paper focuses on the simplest relevant problem (\emph{Minimum Viable Dataset}, MVD) to cut off the obfuscating details and get into the core of what makes neural networks fragile and uninterpretable. Despite much effort, DNNs remain susceptible to adversarial attacks even on MNIST~\cite{schott2018adversarially} and therefore cannot be considered human-aligned even for this simple dataset, see Figure~\ref{fig:mnist_3_5_adv}. That being said, the MVD for this paper is chosen to be the binary subset of MNIST consisting of images of "3" and "5" - as this is one of the most often confused pairs under adversaries~\cite{tian2021analysis} and might be considered harder than the entire MNIST in terms of average adversarial accuracy. For completeness, the preliminary results on the full MNIST are discussed in Appendix~\ref{ap:entire_mnist}.

The paper provides both qualitative and quantitative results. The qualitative analysis focuses on visualising the internal distribution of weights inside layers and how features are matched against the input. For the quantitative results a strong adversarial regime is employed besides computing the standard clean test metrics (computing the clean metrics alone is considered a flawed approach for the reasons discussed earlier).

%% file: theory.tex
In machine learning inputs are represented as tensors. Yet, the standard Euclidean topology in tensor space usually fails to account for small domain-specific variations of features. For example shifting an image by 1 pixel is semantically negligible but usually results in a distant output in $L_2$ metric.

This observation inspires the following definition: 

\paragraph{Definition:}A \emph{semantic feature} $\fpl$ of dimension $s \in \mathbb{N}$ is a tuple $\sfeat$ where:
\begin{itemize}
    \item $f$ - \emph{base feature}, $f \in \mathbb{R}^s$, 
    \item $\mbf{P}$ - \emph{parameter set}, $\mathbf{P} \subset \mathbb{R}^r$ for some $r \in \mathbb{N}$
    \item $\mathbf{L}$ - \emph{locality function} : $\mbf{P} \rightarrow \textrm{Diff}(\mathbb{R}^s)$ where Diff is the family\footnote{informally, $\mbf{L}(\mbf{P})$ should consist of "small" diffeomorphisms.} of differentiable automorphisms of $\mathbb{R}^s$
\end{itemize}

Additionally the feature $\sfeat$ is called \emph{differentiable} iff $\mathbf{L}$ is differentiable (on some neighbourhood of $\mbf{P}$). \hfill \break

Intuitively, semantic feature consists of a base vector $f$ and a set of it's "small" variations $f_p = \mathbf{L}(p)(f)$ for all $p \in \mathbf{P}$; these variations may be called \emph{poses} after~\cite{sabour2017dynamic}. Thus $\sfeat$ is \emph{locality-sensitive} in the adequate topology capturing semantics of the domain. The differentiability of $\mathbf{L}(p)$ and of $\mathbf{L}$ itself allow for gradient updates of $f$ and $\mathbf{P}$ respectively. The intention is to learn both $f$ and $\mathbf{P}$ while defining $\mathbf{L}$ explicitly as an inductive bias\footnote{by the "no free lunch" theorem a general theory of learning systems has to adequately account for the inductive bias.} for the given modality.

\subsection{Examples of semantic features}

To better understand the definition consider the following examples of semantic features:

\begin{enumerate}
    \item \textbf{real-valued} $\fpl$
    \begin{itemize}
        \item $f \in \mathbb{R}$
        \item $\mathbf{P}$ - a real interval $[p_{min}, p_{max}]$ for $p_{min} <= 0 <= p_{max}$
        \item $\mathbf{L}$ - function that maps a real number $p$ to the function $g \mapsto g + p$ for $g \in \mathbb{R}$
    \end{itemize}
    \item \textbf{convolutional} $\fpl$
    \begin{itemize}
        \item $f$ - square 2D convolutional kernel of size $(c_{out}, c_{in}, k, k)$
        \item $\mathbf{P}$ - set of $2\times2$ rotation matrices
        \item $\mathbf{L}$ - function that maps a rotation matrix to the respective rotation of a 2D kernel\footnote{can be made differentiable as in~\cite{eriba2019kornia, jaderberg2016spatial}.\label{fn:kornia}}
    \end{itemize}
    \item \textbf{affine} $\fpl$
    \begin{itemize}
        \item $f$ - vector representing a small 2D spatial feature (e.g. image patch containing a short line)
        \item $\mathbf{P}$ - set of $2\times3$ invertible augmented matrices\footnote{preferably close to the identity matrix.}
        \item $\mathbf{L}$ - function that maps a $2\times3$ matrix to the respective 2D affine transformation of images\footref{fn:kornia}
    \end{itemize}
    \item \textbf{xor} $\fpl$
    \begin{itemize}
        \item $f$ - single-entry vector of dimension $n \in \mathbb{N}$
        \item $\mathbf{P} \subset \{0, 1, \ldots, n-1\}$
        \item $\mathbf{L}$ - function that maps a natural number $p$ to the function that rolls elements of $g \in \mbb{R}^n$ by $p$ coordinates to the right
    \end{itemize}
    \item \textbf{logical} $\fpl$
    \begin{itemize}
        \item $f$ - concatenation of $k$ single-entry vectors and some dense vectors in-between
        \item $\mathbf{P} \subset \{0, 1, \ldots, n_0-1\} \times \ldots \times \{0, 1, \ldots, n_{k-1}-1\}$
        \item $\mathbf{L}$ - function that maps $p \in \mathbf{P}$ to the respective rolls of single-entry vectors (leaving the dense vectors intact)
    \end{itemize}
\end{enumerate}

\subsection{Matching semantic features}\label{sec:SFmatch}

The role of semantic features is to be matched against the input to uncover the general structure of the data. Suppose that we have already defined a function \emph{match}: $(\mathbb{R}^c, \mathbb{R}^s) \rightarrow \mbb{R}$ that measures an extent to which datapoint $d \in \mbb{R}^c$ contains a feature $g \in \mbb{R}^s$. Now let's define

\begin{equation}
\mathrm{SFmatch}(d, \sfeat) = \max_{p \in \mbf{P}} \mathrm{match}(d, \mbf{L}(p)(f))
\label{eq:sfmatch}
\end{equation}

If $\mbf{L}$ captures domain topology adequately then we may think of matching $\fpl$ as a form of local inhibition along $\mbf{L}_{\mbf{P}}(f) = \{L(p)(f), p \in P\}$ and in this sense semantic feature defines a \textbf{XOR} gate over $\mbf{L}_{\mbf{P}}(f)$. On the other hand, every $f_p \in \mbf{L}_{\mbf{P}}(f)$ can be viewed as a conjunction (\textbf{AND} gate) of its non-zero coordinates\footnote{i.e. $f_p$ is a conjunction of lower-level features.}. Thus, semantic features turn out to be a natural way of expressing logical claims about the classical world - in the language that can be used by neural networks. This is particularly apparent in the context of logical semantic features, which can express logical sentences explicitly, e.g. sentence $(A \vee B) \wedge \neg C \wedge \neg D$ can be expressed by a logical $\fpl$ where:
\begin{itemize}
    \item $f = [1, 0, -1, -1]$; \hspace{1 cm} $\mbf{P} = \{0, 1\}$; \hspace{1 cm} $\mbf{L}(0)(f) = [1, 0, -1, -1]$ and $\mbf{L}(1)(f) = [0, 1, -1, -1]$.
\end{itemize}

It remains to characterise the \emph{match} function. Usually it can be defined as the scalar product $d \cdot emb(g)$ where $emb : \mbb{R}^s \rightarrow \mbb{R}^c$ is some natural embedding of the feature space to the input space. In particular for our previous examples lets set the following:
\begin{itemize}
    \item for real-valued $\fpl$: $\mathrm{match}(d, g) = \mathrm{int}(d == g)$ where $d, g \in \mbb{R}$
    \item for convolutional\footnote{an alternative definition would treat every $g^i$ as a separate semantic feature; we stick to the \emph{max} option for simplicity.} $\fpl$: $\mathrm{match}(d, g) = \max_{0 <= i < c_{out}} (d * \frac{g^i}{||g^i||_2})$ where $g$ is a 2D convolutional kernel of shape $(c_{out}, c_{in}, k, k)$, $g^i$ is the $i$-th convolutional filter of $g$, $d$ is an image patch of shape $(c_{in}, k, k)$
    \item for affine $\fpl$: $\mathrm{match}(d, g) = d \cdot \frac{g}{||g||_2}$ where $g$ is a 2D image of shape $(c_{in}, k, k)$ and $d$ is a 2D image patch of the same shape (treated as vectors in $\mbb{R}^c$)
    \item for logical $\fpl$: $\mathrm{match}(d, g) = d \cdot g$ for $g, d \in \mbb{R}^c$
\end{itemize}

The occasional $L_2$ normalization in the above examples is to avoid favouring norm-expanding $\mbf{L}(p)$.

%% file: architecture.tex
In this section we will use semantic features to build a PoC white box neural network model that classifies MNIST images of "3" and "5" in a transparent and adversarially robust way. We will build the network layer by layer arguing carefully for every design choice. Note that the network construction is not an abstract process and must reflect the core characteristics of the chosen dataset. There is no free lunch, if we want to capture the semantics of the dataset we need to encode it in the architecture - more or less explicitly. The framework of semantic features allows to do this in a pretty natural way. Note that a single neural network layer will consist of many parallel semantic features of the same kind.

\begin{figure}
  \centering
  \includegraphics[scale=0.3]{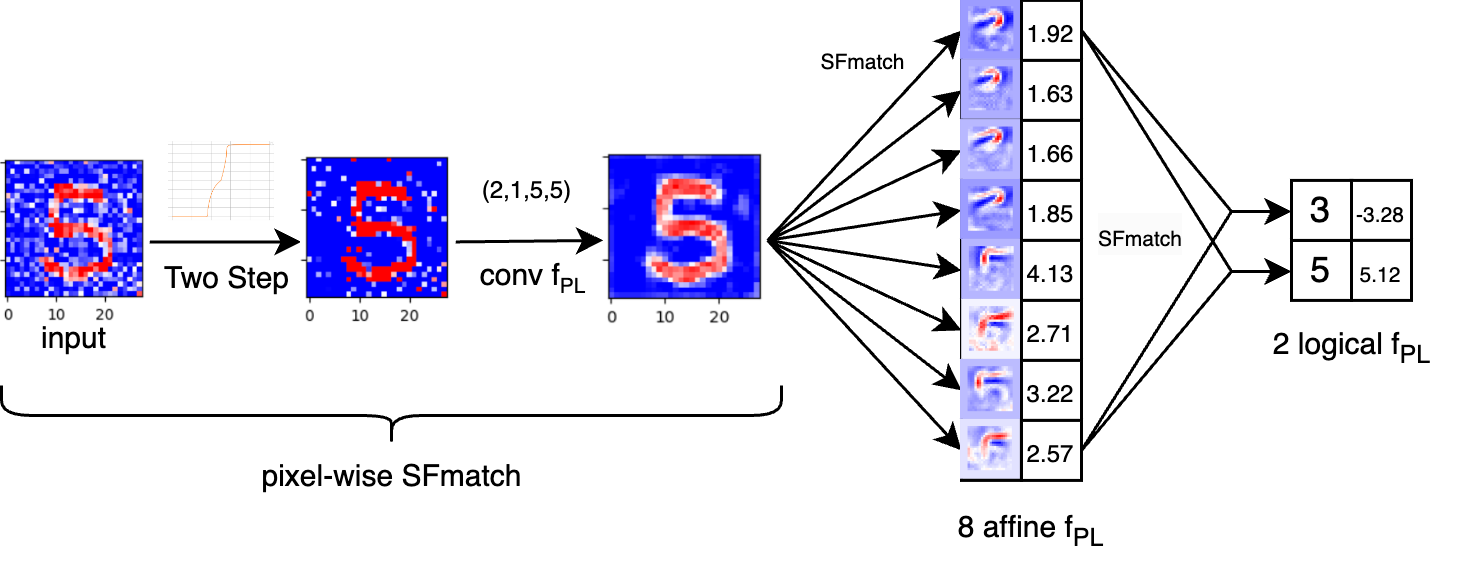}
  \caption{Architecture of the PoC white box neural network. The first two layers consist of semantic features that operate per-pixel - the first layer takes into account only the pixel's value, while the second examines its $5\times5$ neighborhood to determine if the pixel lays on a "bright line". The first two layers retain the shape of the input image. The third layer comprises of 8 affine $\fpl$ as visualized in Figure~\ref{fig:SFmatch}. The final layer consists of two logical $\fpl$: intuitively, the first one checks whether at least \textbf{one} affine $\fpl$ corresponding to "3" is active and \textbf{none} of the affine $\fpl$ corresponding to "5" are active; the second logical $\fpl$ works in the opposite way. Section~\ref{sec:architecture} describes the architecture in more detail.}
  \label{fig:architecture}
\end{figure}

The network consists of the following 4 layers stacked sequentially. For simplicity we don't add bias to any of those layers. Te resulting model has around $5$K parameters. The architecture is visualized in the Figure~\ref{fig:architecture}.

\subsection{Two Step Layer}

MNIST datapoint is a $28\times 28$ grayscale image. Its basic building block is a pixel $x \in [0, 1]$. Despite being allowed to assume any value between $0$ and $1$ it is actually (semantically) a ternary object - it's either ON, OFF or MEH\footnote{the MEH state is an undecided state in-between.}. Therefore the semantic space should squash the $[0, 1]$ interval into those 3 values.

A real-valued $\fpl$ with \emph{match} and $\mbf{L}$ defined as in the previous section identifies numbers in certain interval around $f$. This means that a layer consisting of multiple real-valued $\fpl$, provided that the relevant intervals don't overlap, can be represented as a single locally-constant real-valued function. This allows us to simplify the implementation of such layer and instead of learning the constrained parameters of multiple $\fpl$ we can learn a single parameterised real-valued function that serves as an entire layer. In order to group pixel intensities into 3 abstract values we can draw inspiration from the $\mathrm{softsign}(x) = \frac{x}{1 + x}$ function or rather from it's parameterised version $\mathrm{softsign}(x, t, s) = \frac{x - t}{s + x}$ and ``glue'' two of such step functions together to obtain (Parameterised) Two Step function shown in Figure~\ref{fig:twostep}. The exact implementation of the Two Step function is provided in Appendix~\ref{ap:twostep}. We initialize the layer setting init\_scales\_div=10.

\subsection{Convolutional Semantic Layer}

A pixel-wise function is not enough to classify pixel as ON or OFF - the semantics of our MVD require a pixel to exist in a sufficiently bright region to be considered as ON. Therefore the next layer will consist of a \textbf{single} convolutional semantic feature $\sfeat$:
\begin{itemize}
    \item $f$ - 2D kernel of shape $(2, 1, 5, 5)$ (concatenation of $(g^0, g^1)$, both of shape $(1, 1, 5, 5)$)
    \item $\mbf{P}$ - fixed set of $k$ distinct rotations by a multiple of the $\frac{360^{\circ}}{k}$ angle, for $k = 32$
    \item $\mbf{L}$ - as defined earlier
\end{itemize}

The layer performs $\mathrm{SFmatch}(d(x, y), \fpl)$ with every pixel $(x, y)$ of image $d$. This means that it matches the two rotated filters $g^0$ and $g^1$ with the $5\times5$ neighbourhood of $(x, y)$ and takes the maximum across all the matches\footnote{a single number, not a pair of numbers, as in the chosen definition of match for convolutional $\fpl$.}. Intuitively, this layer checks if pixels have adequately structured neighbourhoods.

The layer is followed by ReLU activation. We initialize $g^0$ and $g^1$ each as the identity kernel of shape $(1, 1, 5, 5)$ plus Gaussian noise $\sim\mathcal{N}(0, 0.1)$.

\subsection{Affine Layer}\label{sec:affine_layer}

The basic semantic building blocks of MNIST digits are fragments of various lines together with their rough locations in the image (same line at the top and the bottom often has different semantics). In short they are shapes at locations. The semantic identity of shape is not affected by small\footnote{close to the identity augmented matrix.} affine transformations. The next shape-extracting layer will therefore consist of \textbf{8} affine semantic features $\sfeat$ defined as follows:
\begin{itemize}
    \item $f$ - 2D image patch of shape $(1, 20, 20)$
    \item $\mbf{P}$ - set of 32 affine $2\times3$ augmented matrices
    \item $\mbf{L}$ - as defined earlier
\end{itemize}

For simplicity every $\fpl$ in this layer is zero-padded evenly on every side and identified with the resulting $28\times 28$ image. Intuitively, this matches the image against a localized shape in a way that is robust to small affine perturbations of the shape.

The layer is followed by ReLU activation. We initialize affine matrices as $2\times 2$ identity matrix plus Gaussian noise $\sim\mathcal{N}(0, 0.01)$ concatenated with $1\times 2$ matrix filled with Gaussian noise $\sim\mathcal{N}(0, 1)$.

\subsection{Logical Layer}

After Affine Layer has extracted predictive shapes it remains to encode logical claims such as "number 3 is A or B and not C and not D" etc. Since the previous layer is learnable, we can enforce a preferable structure on the input neurons to the Logical Layer. This layer will consist of \textbf{2} logical $\fpl$ (one for every label) defined as follows:
\begin{itemize}
    \item $f$ - vector of dimension 8
    \item $\mbf{P} = \{0, 1, 2, 3\}$
    \item $\mbf{L}$ - as defined earlier
\end{itemize}

We initialize  $f^0 = [\frac{4}{10}, 0, 0, 0, -\frac{1}{10}, -\frac{1}{10}, -\frac{1}{10}, -\frac{1}{10}]$ and  $f^1=[-\frac{1}{10}, -\frac{1}{10}, -\frac{1}{10}, -\frac{1}{10}, \frac{4}{10}, 0, 0, 0, ]$. The exact initialization values are less important then their sign and relative magnitude, however they should not be too large as we will train the network using CrossEntropyLoss.

%% file: results.tex
\subsection{Training}

The model was trained to optimize the CrossEntropyLoss for 15 epochs with Adam optimizer, learning rate $3e{-3}$, weight decay $3e{-6}$. The \textbf{only augmentation was Gaussian noise} $\sim\mathcal{N}(0.05, 0.25)$ applied with $0.7$ probability\footnote{mean is slightly positive because of the sparsity of data - subtracting pixel value is on average more harmful than adding it.}. Those parameters were chosen "by hand" after few rounds of experimentation. The results seem pretty robust to the selection of hyperparameters as long as they are within reasonable bounds. Training a single epoch on a single CPU takes around 9s.

\begin{figure}
    \begin{minipage}[t]{0.5\textwidth}
      \centering
      Reliability curve \\
      \vspace{0.5cm}
      \includegraphics[scale=0.5]{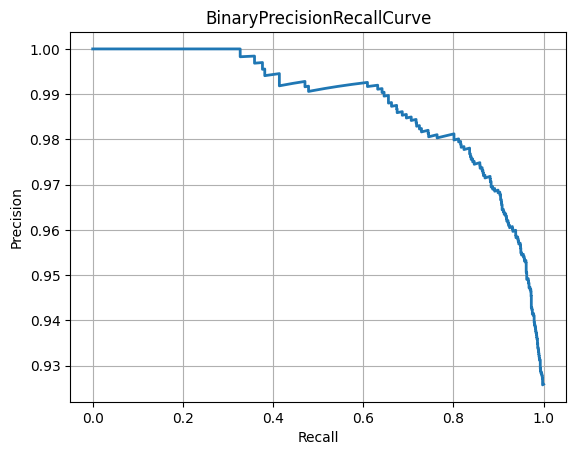}
    \end{minipage}
    \begin{minipage}[t]{0.5\textwidth}
      \centering 
      Learning curve \\
      \vspace{0.1cm}
      \includegraphics[scale=0.5]{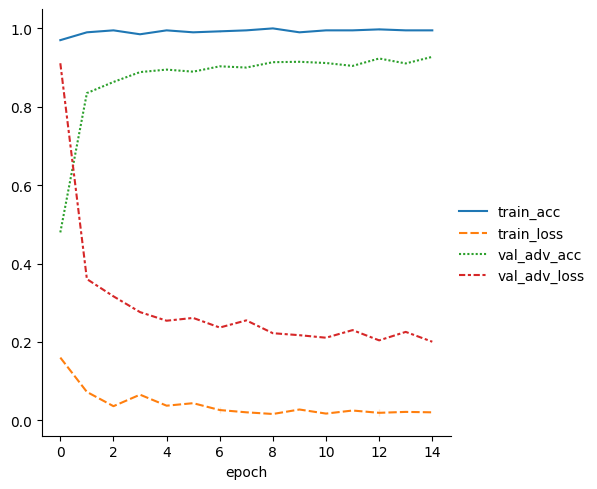}  
    \end{minipage}

  \caption{(left) Reliability test-time curve (see Section~\ref{sec:quantitative_results} for details). (right) Learning curve. Both test and validation metrics presented here are computed under the classic 40-step PGD Attack.}
  \label{fig:curves}
\end{figure}

\subsection{Quantitative results}\label{sec:quantitative_results}

The model achieves \textasciitilde92\% adversarial test accuracy under AutoAttack~\cite{croce2020reliable} (with the default parameters: norm='inf', eps=0.3, step\_size=0.1). The results are almost identical for the much faster 40-step PGD Attack with the same parameters. Adversarial samples are usually misclassified with low confidence and therefore the \emph{reliability curve} is computed for the PGD-attacked test images, i.e. a precision-recall curve where the positive class is defined as the set of correctly classified samples (this means that the precision for 100\% recall is exactly equal to the accuracy). It turns out that for 80\% adversarial recall \textbf{model achieves human-level 98\% adversarial precision} (Figure~\ref{fig:curves}). This means that in practice human intervention would be required only for the \textbf{easily detected} \textasciitilde20\% of \textbf{adversarial} samples to achieve \textbf{human-level reliability} of the entire system (and most real-life samples are \textbf{not} adversarial).

The clean test accuracy is \textasciitilde99.5\%.

\subsection{Layer inspection}

\subsubsection{Two Step Layer}

\begin{figure}
  \centering
  \includegraphics[scale=0.5]{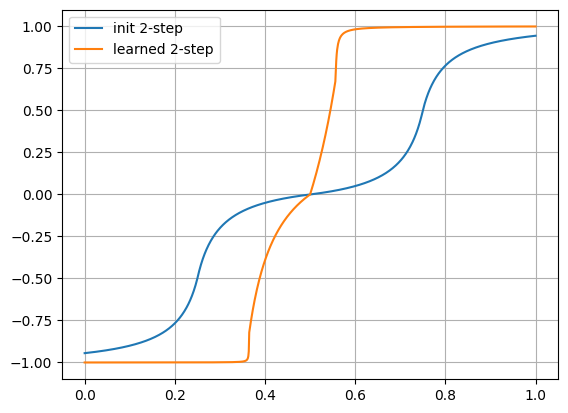}
  \caption{Two Step layer - initial and learned. The visible smooth thresholding essentially means that the layer has learned the permissible perturbations (i.e. intervals) of \emph{real} semantic features corresponding to the state of being OFF (number -1) and ON (number 1).}
  \label{fig:twostep}
\end{figure}

Figure~\ref{fig:twostep} shows the initial and learned shapes of the Two Step function. The function smoothly thresholds input at $0.365$ and then at $0.556$ which means that the interval $[0, 0.365]$ is treated as OFF, $[0.365, 0.556]$ as MEH and $[0.556, 1]$ as ON. Bear in mind that this is not the final decision regarding the pixel state - it's just the input to the next layer.

\subsubsection{Convolutional Semantic Layer}

\begin{figure}
  \centering
  \includegraphics[scale=0.26]{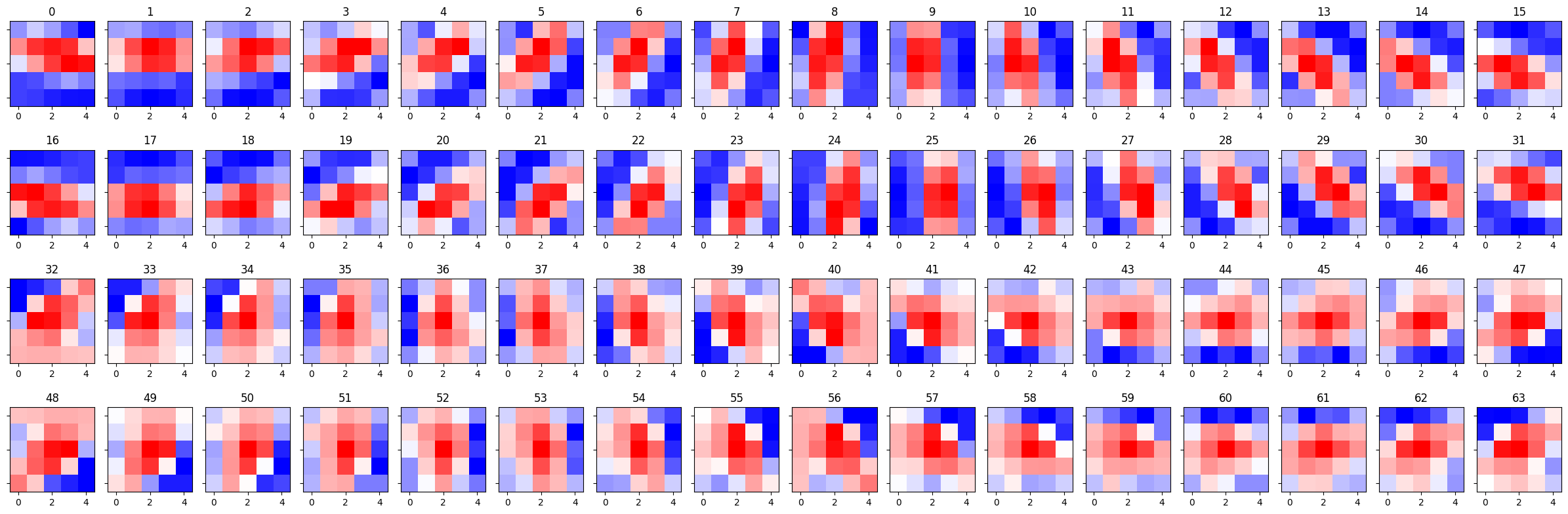}
  \caption{Learned convolutional kernels in 32 predefined poses. Convolutional Semantic Layer acts pixel-wise and matches (convolves) these rotated kernels against the $5\times5$ neighbourhood of a given pixel, returning the maximum matching value. Intuitively, this corresponds to the degree of pixel laying on a "bright line".}
  \label{fig:conv}
\end{figure}

Quick inspection of the Figure~\ref{fig:conv} shows that the Convolutional Semantic Layer has indeed learned meaningful filters that check if the central pixel lays inside an adequately structured region.

\subsubsection{Affine Layer}\label{sec:affine_results}

\begin{figure}
  \centering
  \includegraphics[scale=0.53]{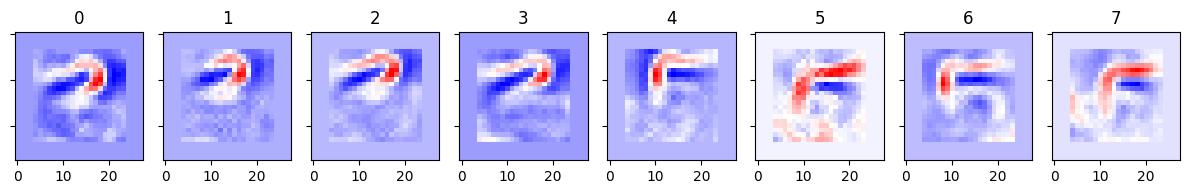}
  \caption{Learned affine base features. The weight-sharing strategy imposed by the affine \emph{locality function} results in easily interpretable features.}
  \label{fig:affine}
\end{figure}

Figure~\ref{fig:affine} shows the learned affine base features. Appendix~\ref{ap:affine_poses} shows the 8 learned affine features in all of the 32 learned feature-specific poses. It's hard to imagine substantially more meaningful features.

\subsubsection{Logical Layer}

The two learned logical base features are as follows:
\begin{itemize}
    \item $f^0 = [1.0700, 0.0000, 0.0000, 0.0000, -0.7800, -0.7100, -0.8000, -0.8100]$
    \item $f^1 = [-0.8100, -0.7600, -0.8600, -0.8000, 1.0500, 0.0000, 0.0000, 0.0000]$
\end{itemize}

This is exactly as expected - $\fpl^0$ defines number "3" as an entity having large SFmatch with \textbf{any} of the first 4 affine semantic features and small SFmatch with \textbf{all} of the remaining 4 affine semantic features. For $\fpl^1$ its the other way round.

\begin{figure}
  \centering
  \includegraphics[scale=0.26]{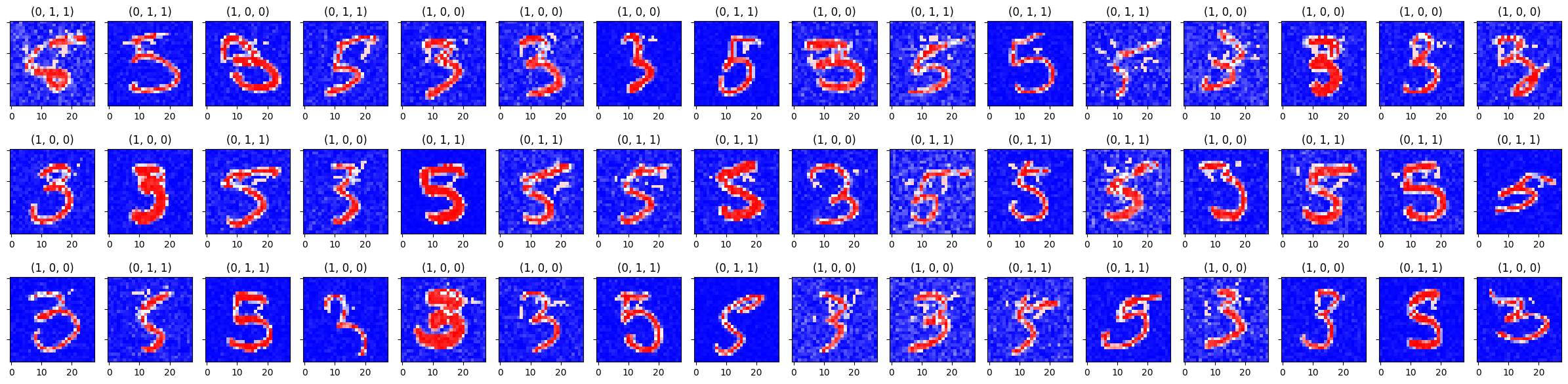}
  \caption{Minimally perturbated images under Boundary Attack~\cite{brendel2018decisionbased} to change model's predictions - for sample test images. The perturbations are often meaningful to humans and introduce semantic ambiguity. Note that by definition of the Boundary Attack the model is extremely unconfident on those samples (model confidence at \textasciitilde50\%) and in practice the perturbations required to fool the entire system would have to be much larger.}
  \label{fig:boundary}
\end{figure}

\begin{figure}
  \centering
  \includegraphics[scale=0.26]{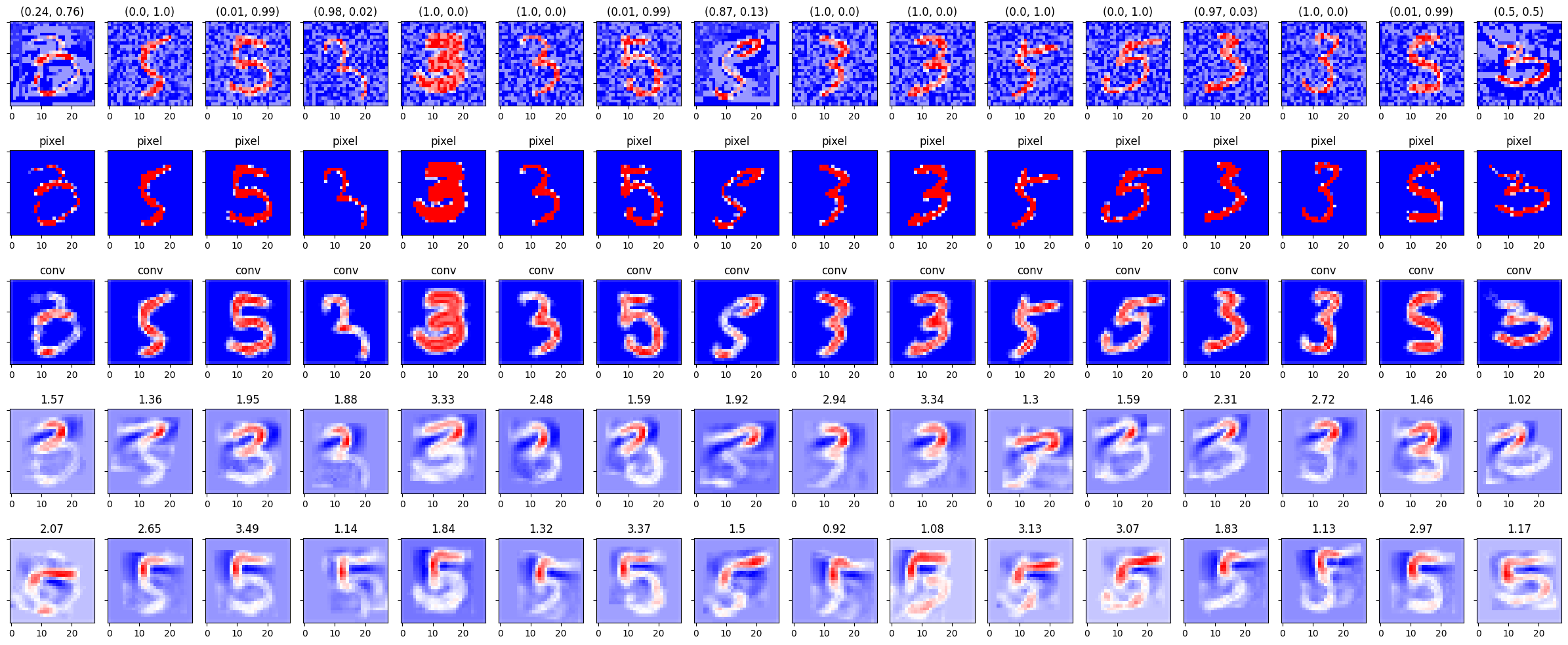}
  \caption{Visualization of test data as "seen" by the first 3 layers. The consecutive rows and labels denote respectively: adversarial input and predicted class probabilities; output of Two Step Layer; output of Convolutional Semantic Layer; top pose of affine features predicting label "3" (with the match value above); same for label "5". The last two rows are overlaid with the middle row to make it easier to see what has been matched. Note that there are 3 misclassified examples (first, 8th and last) and for every one of them \textbf{it's easy to understand why the model made the mistake}: for the first one the model recognizes the middle part of "3" as the top part of "5"; the "5" in the 8th image has specific curve at the top which resembles the top part of "3"; the "3" in the last example has very thin top part which gets removed by the first 2 layers thus creating an ambiguity.}
  \label{fig:layer_vis}
\end{figure}

\subsection{Ablation study}\label{sec:ablation}

Most of the heavy lifting regarding pixel-intensity alignment is done by the Two Step Layer which lifts adversarial robustness from 0\% to \textasciitilde86\%. The Convolutional Semantic Layer stabilizes the training and boosts adversarial robustness by further \textasciitilde6\%pt.

Interestingly the number of affine semantic features can be reduced from \textbf{8} to \textbf{2} with the cost of just \textasciitilde2\%pt adversarial accuracy. Logical Layer can be replaced by a simple linear probe, however it needs a proper initialization to avoid the risk of dead neurons (which is high because of the exceedingly small amount of neurons).

The Gaussian noise augmentation is vital for the model to experience pixel intensity variations during training (boosts adversarial robustness by \textasciitilde62\%pt).

Despite being primarily designed to withstand the $L_{\infty}$ attack the model has considerable robustness against the standard $L_2$ AutoPGD attack (eps=1.5, eps\_step=0.1) with \textasciitilde89\% adversarial accuracy.

%% file: discussion.tex
We've built a white box neural network that is naturally robust, reliable and inherently interpretable. Thanks to the notion of semantic features the construction of the model was as straightforward as writing a fully explainable procedural program and so are the inner workings of the network. 

\subsection{Defending against adversaries}

Affine Layer is the core component of the model and the main reason for its interpretability. However if applied directly to the input it is easily fooled by attacks. \textbf{Its white box nature makes it easy to understand where this vulnerability comes from and to fix it}: quick inspection of the mismatched poses as in the Figure~\ref{fig:layer_vis} reveals that it is predominantly due to the large attack budget of $0.3$ per pixel (which is $30\%$ of the entire allowed pixel intensity range) combined with the large linear regions of the function computed by this layer (due to small amount of poses per feature). Human eye appears to ignore such considerate changes in pixel intensities which means that the naive linear representation of pixel brightness is not semantically adequate. This is exactly how (and why) the first two layers of the model were designed - to fix the inherent vulnerability of otherwise perfectly interpretable affine features. These layers are themselves compatible with the framework of semantic features which is an early indication of a large potential of the notion.

\subsection{Limitations}

Semantic features require \emph{locality engineering} as they rely on the adequate domain-specific definition of the \emph{locality function}. Therefore, they won't be applicable in contexts where such locality is hard to define. On the other hand, it opens up interesting areas for research.

The Logical Layer has been developed as an attempt to formalize the idea of how the generic hidden layer activation pattern should look like; however, it might turn out to be prone to the "no free lunch" theorem (e.g. on MNIST the properly-initialized linear probe seems to be sufficient).

Despite the generality of the notion and the ideas given in Section~\ref{sec:further} there is a risk that the white box qualities of semantic features will be difficult to achieve for harder datasets. The preliminary results on CIFAR10 seem promising but also reveal the need for careful engineering of the locality function, see Section~\ref{sec:further}.

%% file: further_research.tex
Notice the analogy between affine semantic features and the SIFT~\cite{sift} algorithm - the learnable pose sampling may be interpreted as finding Keypoints and their Descriptors, while the initial 2 layers of the PoC network essentially perform input normalization. This might be a helpful analogy for designing more advanced semantic features that sample across varying scales, sharpness or illumination levels.

There might be a concern that adding more layers to the network would reduce its interpretability. However, the role of additional layers would be to widen the receptive field of the network. The Affine Layer defined earlier matches \emph{shapes-at-locations} of center-aligned objects. Adding additional layer would involve a sliding window approach (or Keypoint Detection similar to that in SIFT) to match center-aligned objects inside the window and integrate this with the approximate global location of the window; then the "Global" Affine Layer would match \emph{objects-at-locations} forming a scene (a higher-level object). This would in fact be similar to applying Affine Layer to the "zoomed-out" input.

The ablations in Section~\ref{sec:ablation} suggest that adversarial vulnerability of DNNs might be mostly due to the fragility of the initial layers. Therefore, another direction forward would be to re-examine the known approaches to XAI and try to "bug fix" their encoder networks in similar fashion.

Other ideas for further work on semantic features include:
\begin{itemize}
    \item use semantic features in self-supervised setting to obtain interpretable dimensionality reduction
    \item study non-affine perturbations of spatial features, e.g. projective perturbations
    \item find adequate semantic topology for color space
    \item semantic features for sound (inspiration for semantic invariants could be drawn from the music theory)
    \item semantic features for text, e.g. semantic invariants of permutations of words/sentences capturing \emph{thoughts}
    \item study more complex logical semantic features in more elaborate scenarios
    \item figure out a way to make logical semantic features differentiable, i.e. make $\mbf{P}$ learnable
\end{itemize}

%% file: acknowledgments.tex
This is work has been done as an independent self-funded research.

However it needs to be noted that I had the opportunity to do preliminary investigation on adversarial robustness as a half-time activity during my employment at MIM Solutions. Special thanks go to Piotr Sankowski, who introduced me to the issue of adversarial vulnerability of neural networks, and both to Piotr Sankowski and Piotr Wygocki who - as my former employers - decided to trust me with that research as a half-time part of my contract. It was precisely thanks to this opportunity that I developed a strong intuition that modern neural network architectures are inherently flawed and need to be fundamentally re-examined.

I thank Piotr Wodziński, Adam Szummer, Tomasz Steifer, Dmytro Mishkin, Andrzej Pacuk, Bartosz Zieliński, Hubert Baniecki and members of the reddit community for the great feedback on the initial versions of this paper.

%% file: appendix.tex
\section{Implementation of Two Step Layer}\label{ap:twostep}

\begin{lstlisting}[language=Python]
class TwoStepFunction(Module):
    """Simplified real-valued SFLayer for inputs in [0, 1]"""
    def init_weights(self):
        self.a0 = nn.Parameter(torch.tensor(0.5))
        self.a1 = nn.Parameter(torch.tensor(0.5))
        self.t0 = nn.Parameter(torch.tensor(0.25))
        self.t1 = nn.Parameter(torch.tensor(0.75))

        self.scales = nn.Parameter(
            (torch.ones((4,)) / self.init_scales_div)
        )

    def softsign(self, x, threshold, scale, internal):
        a = self.a1 if threshold > 0 else self.a0

        x = x - threshold
        x = x / (scale + x.abs())

        if internal:
            shift = threshold
            shift = shift / (scale + shift.abs())
            x = x + shift
            x = a * x / shift.abs()

            return x

        div = x.abs().max() / (1 - a)
        x = x / div
        x = x + threshold.sgn() * a

        return x

    def transform(self, x):
        return 2 * x - 1

    def forward(self, x):
        x = self.transform(x)
        t0, t1 = self.transform(self.t0), self.transform(self.t1)

        params = [
            (t0, self.scales[0], False),
            (t0, self.scales[1], True),
            (t1, self.scales[2], True),
            (t1, self.scales[3], False),
        ]
        xs = [self.softsign(x, *p) for p in params]

        masks = [
            (x < t0).float(),
            (t0 <= x).float() * (x < 0.0).float(),
            (0.0 <= x).float() * (x < t1).float(),
            (t1 <= x).float(),
        ]
        x = sum(m * xx for (m, xx) in zip(masks, xs))
        return x
\end{lstlisting}
\newpage

\section{Learned affine features with learned poses}\label{ap:affine_poses}
\begin{figure}[!b]
  \centering
  \includegraphics[scale=0.26]{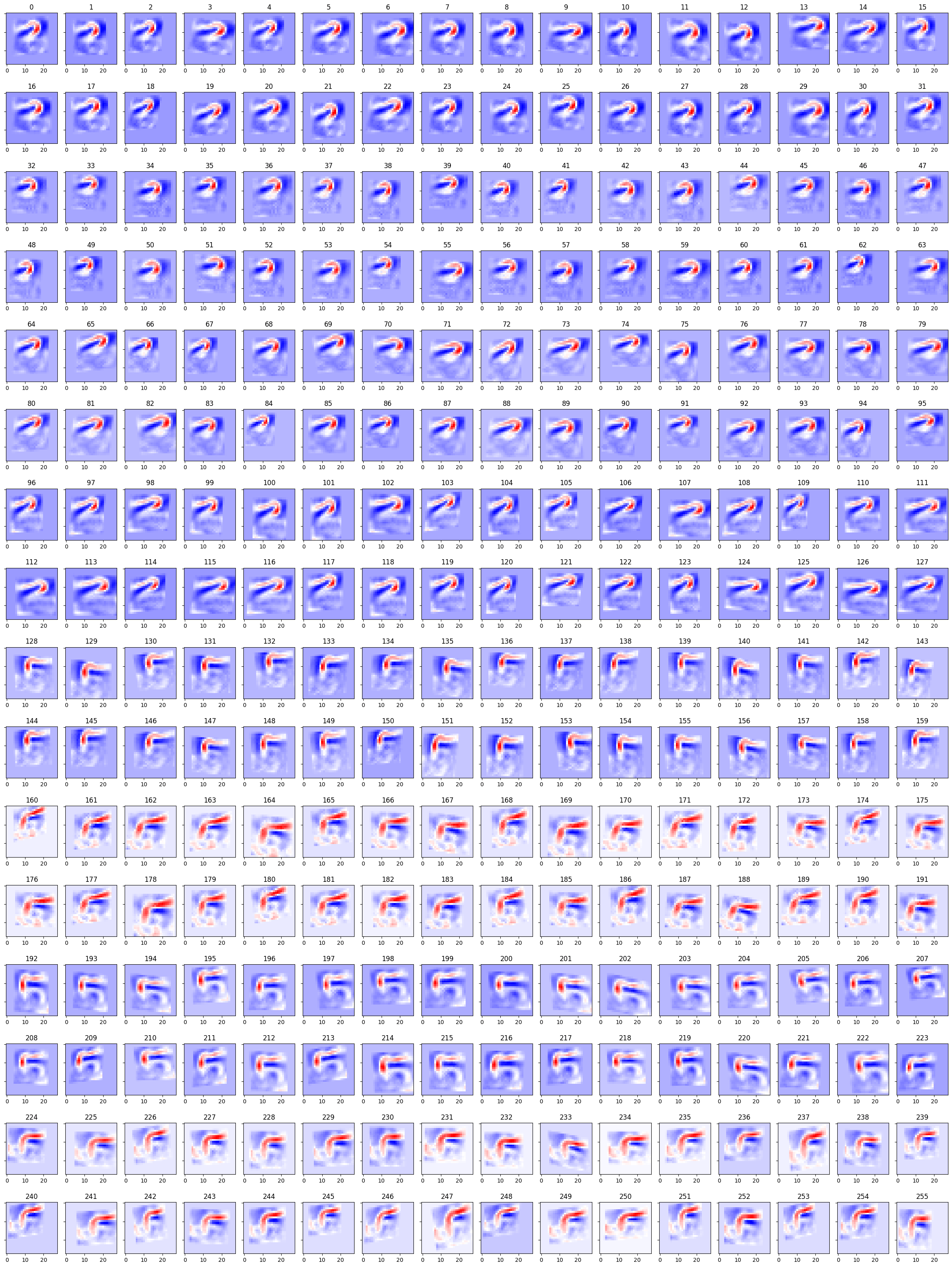}
  \label{fig:affine_poses}
\end{figure}
\newpage

\section{Preliminary results for the entire MNIST}\label{ap:entire_mnist}

Let's review the results for a straightforward adaptation of previously defined model for the full MNIST:

\begin{itemize}
    \item Affine Layer is expanded to \textbf{50} affine semantic features;
    \item Logical Layer is replaced by a linear probe with 10 output classes;
    \item Size of affine features is reduced to $14\times14$ patches;
    \item Affine \emph{base} features are equipped with learnable $(x, y)$ per-feature location;
\end{itemize}

Such model has \textasciitilde20K parameters, achieves \textasciitilde98\% standard and \textasciitilde80\% adversarial accuracy (AutoAttack) and can still be considered \emph{white box} as shown in the following figures:

\begin{figure}[!b]
  \centering
  \includegraphics[scale=0.4]{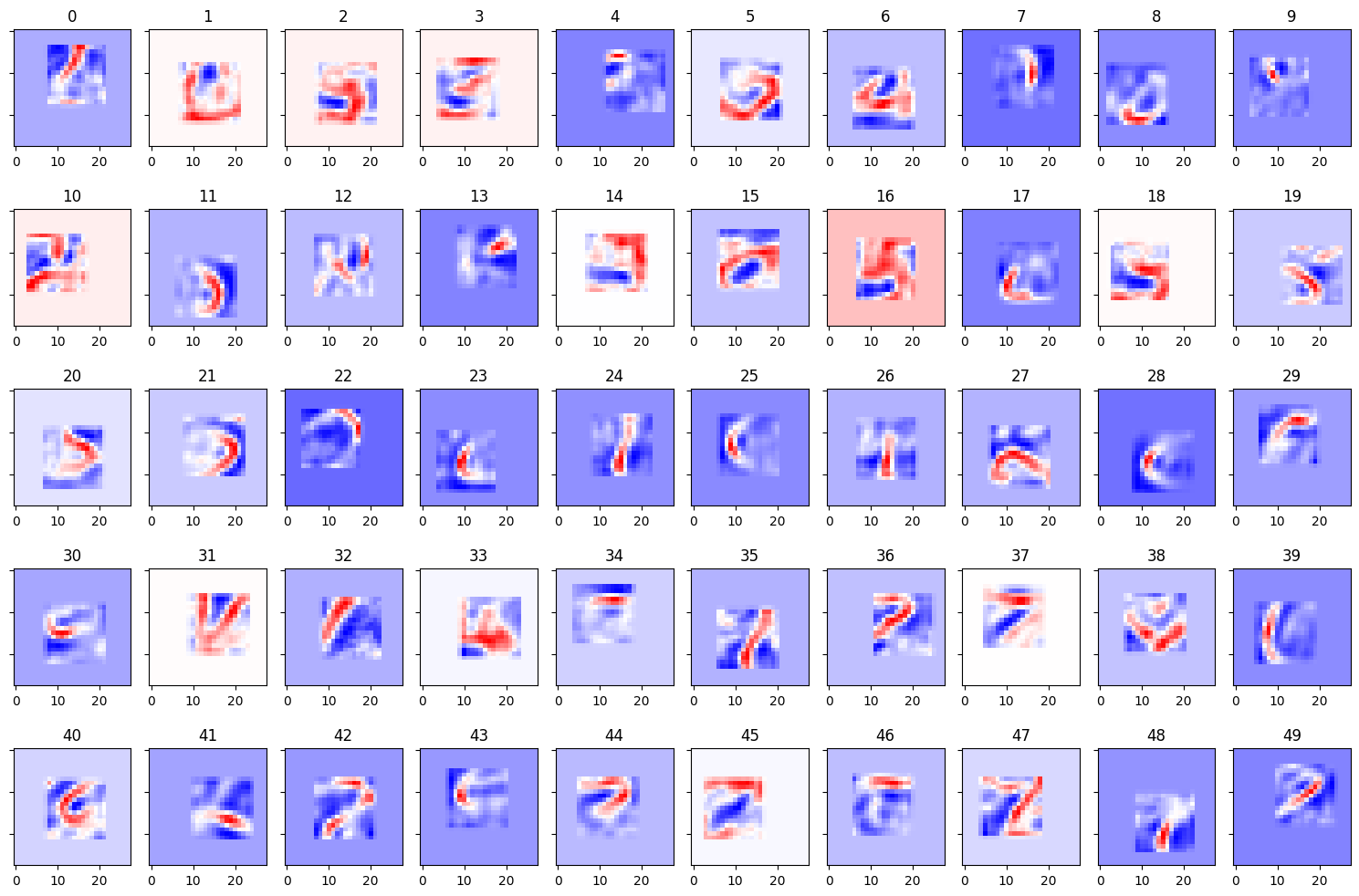}
  \caption{Affine base features learned on full MNIST. Due to the learnable feature-specific $(x, y)$-location these can be considered \emph{shapes-at-locations}.}
  \label{fig:full_affine}
\end{figure}

\begin{figure}[!b]
  \centering
  \includegraphics[scale=0.4]{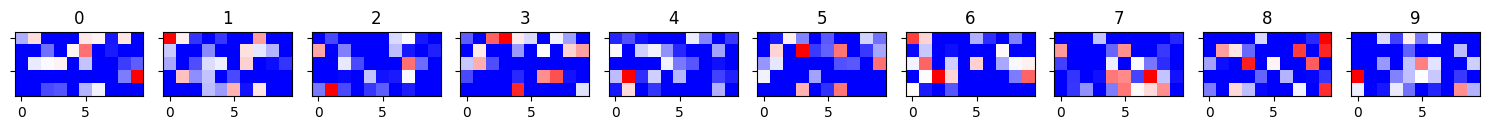}
  \caption{Masked positive weights of linear probe - per output class (here the dark-blue squares are zeros). There are only few dark red neurons per class which means that the digit representation can be considered sparse.}
  \label{fig:positive_probe}
\end{figure}

\begin{figure}[!b]
  \centering
  \includegraphics[scale=0.4]{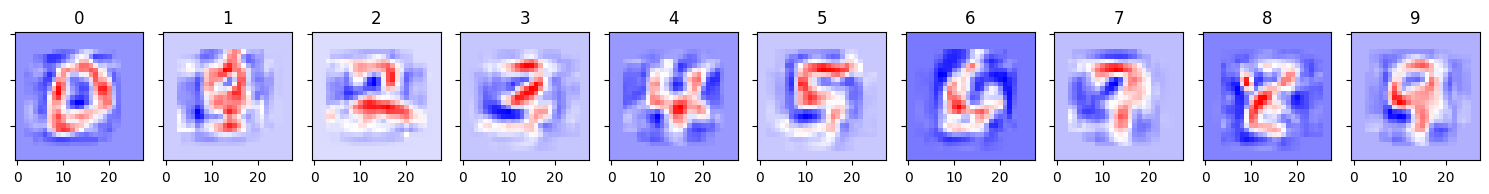}
  \caption{Sum of base affine features weighted by positive weights of linear probe - per output class. The similarity to actual digits means that the model decomposes the digits into human-aligned \emph{shapes-at-locations}.}
  \label{fig:numbers}
\end{figure}